%% file: example_paper.tex
\documentclass{article}

\usepackage{microtype}
\usepackage{graphicx}
\usepackage{subcaption}
\usepackage{booktabs} % for professional tables

\usepackage{hyperref}

\usepackage[preprint]{icml2026}

\usepackage{amsmath}
\usepackage{amssymb}
\usepackage{mathtools}
\usepackage{amsthm}

\usepackage{graphicx}

\usepackage{algorithm}
\usepackage{algorithmic}

\usepackage{amsmath,amsfonts}

\usepackage{amssymb}
\usepackage{array}
\usepackage{subcaption}
\usepackage{caption}
\usepackage{textcomp}
\usepackage{url}
\usepackage{verbatim}
\usepackage{graphicx}
\usepackage{natbib}
\usepackage{xcolor}
\usepackage{soul}
\usepackage{enumitem}
\usepackage{tabularray}

\usepackage{arydshln}
\usepackage{fancyhdr}
\usepackage{varwidth}
\usepackage{arydshln}
\usepackage{multirow}
\usepackage{mathtools}
\usepackage{algorithmic}
\usepackage{algorithm}

\usepackage{datetime2}
\usepackage{pifont}
\usepackage{enumitem}
\usepackage[most]{tcolorbox}
\usepackage{tabularx}
\usepackage{xltabular}

\usepackage{newfloat}
\usepackage{listings}

\usepackage{kotex}
\usepackage{soul}
\usepackage{makecell}

\newcommand{\sys}{\textsf{\small Zero2Text}}

\newcommand{\remove}[1]{}

\usepackage{datetime}
\usepackage[capitalize,noabbrev]{cleveref}

\theoremstyle{plain}

\theoremstyle{definition}

\theoremstyle{remark}

\usepackage[textsize=tiny]{todonotes}

\icmltitlerunning{Zero2Text: Zero-Training Cross-Domain Inversion Attacks on Textual Embeddings}

\begin{document}

\twocolumn[
  %\icmltitle{Token Level Cross-Domain Textual Inversion Attacks}
    \icmltitle{Zero2Text: Zero-Training Cross-Domain Inversion Attacks \\
    on Textual Embeddings}
%via Recursive Cross-Model Alignment and Generation}}


  \icmlsetsymbol{equal}{*}

  \begin{icmlauthorlist}
    \icmlauthor{Doohyun Kim}{equal,kaistgsis}
    \icmlauthor{Donghwa Kang}{equal,kaist}
    \icmlauthor{Kyungjae Lee}{uoscs}
    \icmlauthor{Hyeongboo Baek}{uosai}
    \icmlauthor{Brent ByungHoon Kang}{kaist}
  \end{icmlauthorlist}

  \icmlaffiliation{kaist}{School of Computing, Korea Advanced Institute of Science and Technology (KAIST), Daejeon, Republic of Korea}
  \icmlaffiliation{kaistgsis}{Graduate School of Information Security, Korea Advanced Institute of Science and Technology (KAIST), Daejeon, Republic of Korea}
  \icmlaffiliation{uosai}{Department of Artificial Intelligence, University of Seoul, Seoul, Republic of Korea}
  \icmlaffiliation{uoscs}{Department of Computer Science, University of Seoul, Seoul, Republic of Korea}

  \icmlcorrespondingauthor{Brent ByungHoon Kang}{brentkang@kaist.ac.kr}
  \icmlcorrespondingauthor{Hyeongboo Baek}{hbbeak@uos.ac.kr}

  \icmlkeywords{Machine Learning, ICML}

  \vskip 0.3in
]



\printAffiliationsAndNotice{}  % no special notice (required even if empty)

\input{00abstarct}

\input{01introduction}

\input{02related_works}

\input{04method}

\input{05evaluation}

\input{06conclusion}

%This paper presents work whose goal is to advance the field of machine learning.
%There are many potential societal consequences of our work, none of which we feel must be specifically highlighted here.

\nocite{langley00}

\bibliography{example_paper}
\bibliographystyle{icml2026}

\end{document}

%% file: 00abstarct.tex
\begin{abstract}
The proliferation of retrieval-augmented generation (RAG) has established vector databases as critical infrastructure, yet they introduce severe privacy risks via \textit{embedding inversion attacks}. 
Existing paradigms face a fundamental trade-off: optimization-based methods require computationally prohibitive queries, while alignment-based approaches hinge on the unrealistic assumption of accessible in-domain training data. 
These constraints render them ineffective in strict black-box and cross-domain settings.
To dismantle these barriers, we introduce \sys{}, a novel \textit{training-free} framework based on \textit{recursive online alignment}. 
Unlike methods relying on static datasets, \sys{} synergizes LLM priors with a dynamic ridge regression mechanism to iteratively align generation to the target embedding on-the-fly. 
We further demonstrate that standard defenses, such as differential privacy, fail to effectively mitigate this adaptive threat.
Extensive experiments across diverse benchmarks validate \sys{}; notably, on MS MARCO against the OpenAI victim model, it achieves 1.8$\times$ higher ROUGE-L and 6.4$\times$ higher BLEU-2 scores compared to baselines, recovering sentences from unknown domains without a single leaked data pair.
\end{abstract}

%% file: 01introduction.tex
\section{Introduction}
\label{sec:intro}

\begin{table*}[t]
    \centering
    \caption{Comparison of inversion attack paradigms.}
    \label{tab:methodology_comparison}
    \resizebox{2\columnwidth}{!}{
    \small
    \renewcommand{\arraystretch}{1.2}
    \begin{tabular}{cccc}
        \toprule
        \textbf{Category} & \makecell[c]{\textbf{Direct Inversion Methods}\\ (e.g., Vec2Text, GEIA)} & \makecell[c]{\textbf{Embedder Alignment Methods}\\ (e.g., TEIA, ALGEN)} & \makecell[c]{\textbf{\sys{}} \\ \textbf{(Ours)}} \\
        \toprule
        \textbf{Decoder Training} & \makecell[c]{Required \\ (embedding → text decoder)} & \makecell[c]{Required \\(decoder trained on proxy embedder)} & \makecell[c]{\textbf{Not required}\\ \textbf{(training-free)}} \\ 
        \midrule
        \textbf{Alignment Training} & Not applicable & \makecell[c]{Required\\ (victim ↔ proxy embedder alignment) }& \makecell[c]{\textbf{Not required} \\ \textbf{(online optimization)}} \\
        \midrule
        \textbf{Data Dependency} &  \makecell[c]{Large-scale collections of \\ leaked embedding–text pairs}  & \makecell[c]{Small-scale leaked embedding–text pairs \\ + Large-scale general corpus} & \makecell[c]{\textbf{Few-shot access to } \\ \textbf{the victim model at test time}} \\
        \midrule
        \makecell[c]{\textbf{Domain Assumption} \\ \textbf{(Training ↔ Test sets)}} & Same-domain & \makecell[c]{ Corpus: General domain \\ Alignment: Same-domain} &  \textbf{No dependency} \\
        \midrule
        \textbf{Generalization} & Poor (training domain bias) & Limited (alignment domain bias) & \textbf{Robust (instance-specific)} \\
        \bottomrule
    \end{tabular}
    }
\end{table*}

The proliferation of large language models (LLMs)~\cite{brown2020language, touvron2023llama} has revolutionized natural language processing, yet their inherent limitations, such as hallucinations and a lack of up-to-date knowledge, necessitate external augmentation. 
Consequently, retrieval-augmented generation (RAG)~\cite{lewis2020retrieval} has emerged as a de facto standard, enabling models to access proprietary and temporal data. 
Central to this architecture are vector databases (DBs)~\cite{johnson2019billion}, which store massive amounts of unstructured data—ranging from personal emails to corporate trade secrets—as high-dimensional dense embeddings. 
These semantic representations are widely considered efficient and, implicitly, obfuscated enough to preserve privacy. As a result, vector DBs have become a critical infrastructure in modern AI stacks, handling sensitive information under the assumption that embeddings are non-invertible one-way functions.

\begin{figure}[t!]
    \centering
    \includegraphics[width=1\linewidth]{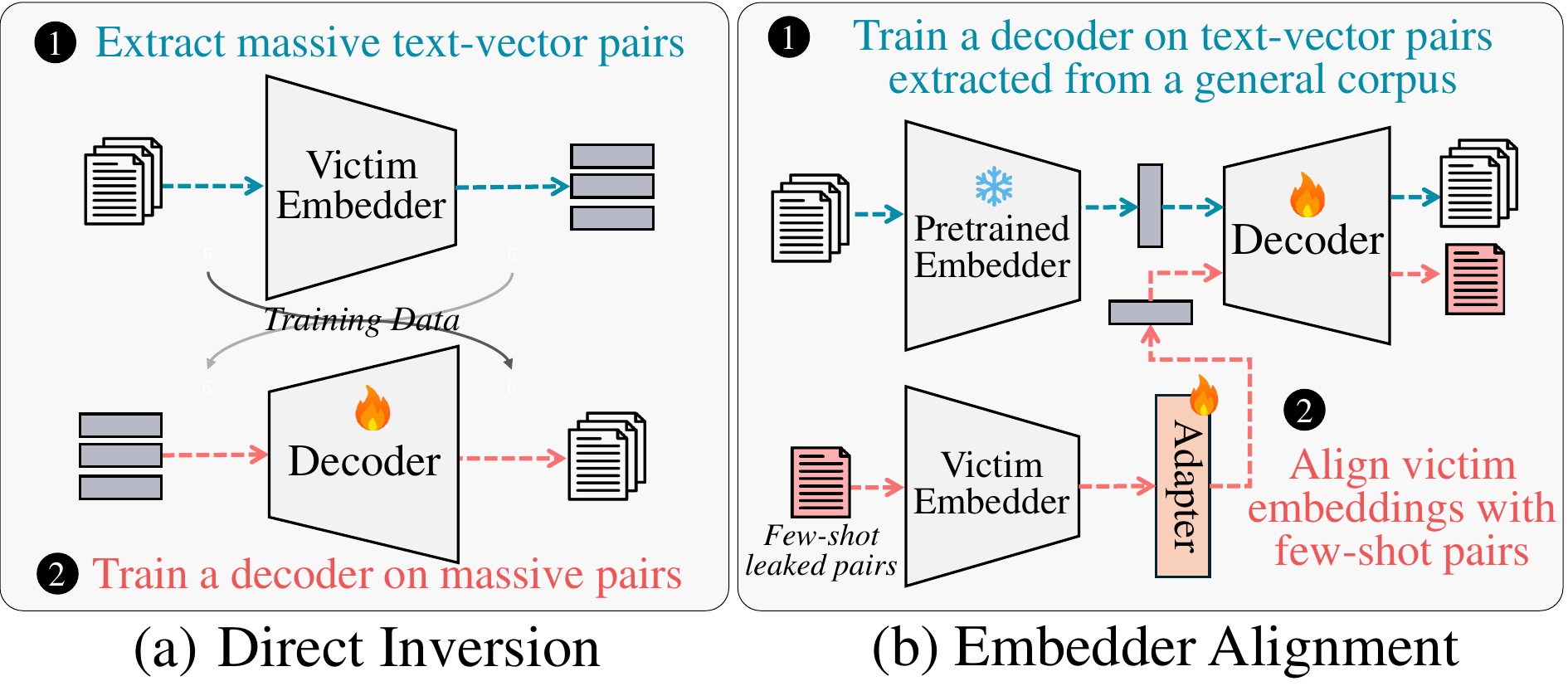}
    \caption{Illustration of prevailing inversion paradigms. (a) Direct inversion methods (e.g., Vec2Text) require extracting massive text-vector pairs (\ding{182}) to train a specialized decoder (\ding{183}). 
(b) Embedder alignment methods (e.g., ALGEN) train a decoder on general corpora (\ding{182}) and rely on few-shot leaked pairs from DB to align the victim embedder (\ding{183}). 
Both paradigms struggle with domain shifts when the target domain is unknown.}
\label{fig:intro}
\end{figure}

However, this assumption of safety is increasingly being challenged. 
Recent studies demonstrate that text embeddings are not cryptographically secure and retain significant semantic information that can be reconstructed. 
This vulnerability exposes a critical attack surface: \textit{embedding inversion attacks}~\cite{song2020information, morris2023text}. 
An adversary with access to DB—whether through a data breach or public API endpoints—can recover the original raw text, leading to severe privacy violations such as the leakage of personally identifiable information (PII)~\cite{carlini2021extracting} or confidential intellectual property.

Current embedding inversion attacks largely fall into two methodological paradigms, as illustrated in Figure~\ref{fig:intro}.
The first category comprises \textit{direct inversion} methods (Figure~\ref{fig:intro}(a)), such as Vec2Text~\cite{morris2023text} and GEIA~\cite{li2023sentence}, which learn a direct mapping from victim embeddings to text. 
While effective under idealized conditions, these approaches rely on a prohibitive assumption: the adversary must extract massive victim embedding–text pairs (\ding{182})—often on the order of millions (e.g., 5M pairs)—to train a decoder (\ding{183}).
Motivated by this limitation, a second line of work adopts \textit{embedder alignment} methods (Figure~\ref{fig:intro}(b)), including ALGEN~\cite{chen2025algen} and TEIA~\cite{huang2024teia}. 
These approaches mitigate data dependency by training a decoder on general corpora (\ding{182}) and using a small number of leaked pairs (typically 1–8K) to align the victim embedder (\ding{183}).
Table~\ref{tab:methodology_comparison} summarizes the trade-offs of these paradigms against our proposed framework.

Despite their differences in data scale, both paradigms share a critical limitation: they implicitly assume that the domain of the target embeddings matches the training or leaked (from DB) alignment data. 
In realistic deployments, embeddings often originate from unknown, heterogeneous, or highly specialized domains. 
Such distributional shifts fundamentally undermine existing threat models, causing significant performance degradation in cross-domain settings.

To address these limitations, we propose \sys{}, a training-free embedding inversion attack based on \textit{search and recovery via online verification}.
Unlike prior methods that rely on large collections of leaked embedding–text pairs or offline alignment, \sys{} requires \textit{no decoder training} and \textit{no leaked alignment data}.
Given a target embedding at inference time, \sys{} leverages a pre-trained LLM as a universal generator and iteratively issues a limited number of online API queries. 
LLM-guided texts are verified by comparing their embeddings to the target embedding, and this similarity signal is recursively fed back to guide subsequent generations. 
Through this instance-specific, on-the-fly process, \sys{} aligns generation to the target embedding without assuming prior knowledge of the embedding domain, enabling robust inversion even under unknown or out-of-distribution settings.

Our contributions are threefold:
\begin{itemize}[leftmargin=*, nosep]
    \item We propose \sys{}, a novel architecture that eliminates reliance on auxiliary datasets. By synergizing recursive LLM generation with targeted online optimization, it achieves state-of-the-art fidelity in strict black-box settings.
    \item We demonstrate that \sys{} significantly outperforms existing baselines (e.g., Vec2Text, ALGEN) in cross-domain generalization, exemplified by achieving 1.8$\times$ higher ROUGE-L and 6.4$\times$ higher BLEU-2 scores on the MS MARCO dataset against the OpenAI model, proving that high-precision inversion is attainable without domain-specific pre-training.
    \item We comprehensively evaluate standard defenses, such as differential privacy and noise injection, exposing their insufficiency against recursive alignment attacks and underscoring the urgent need for stronger embedding security.
\end{itemize}

%% file: 02related_works.tex
\section{Related Work}
\label{sec:related}

\subsection{Inversion Attacks on Embeddings}

The vulnerability of deep learning models to inversion attacks has been extensively documented. 
Early studies by \citet{fredrikson2015model} demonstrated the reconstruction of sensitive inputs from model outputs in image and medical domains. 
In NLP, white-box attacks exploiting gradients have been shown to recover input tokens~\cite{song2020information} and extract user data from general-purpose language models~\cite{pan2020privacy} or federated learning updates~\cite{zhu2019deep, gupta2022recovering}. 
Furthermore, \citet{carlini2021extracting} revealed that LLMs memorize training data, which can be elicited via targeted prompting. 
Recent research has pivoted to black-box settings relevant to embedding APIs. 
\citet{li2023sentence} empirically showed that sentence embeddings leak significant semantic information. \citet{morris2023text} established a strong baseline with Vec2Text, an optimization-based method that iteratively refines inputs to match target embeddings, while \citet{balunovic2022lamp} explored discrete optimization for similar purposes. 
Most recently, \citet{chen2025algen} introduced ALGEN, a generation-based framework leveraging cross-model alignment for few-shot inversion, offering a more query-efficient alternative to optimization-based approaches. 
Unlike prior works relying on auxiliary datasets or static alignment, \sys{} introduces a zero-shot paradigm that eliminates data dependencies by dynamically aligning the embedding space online via a recursive feedback loop.

\begin{figure*}[t]
    \centering
    \includegraphics[width=\linewidth]{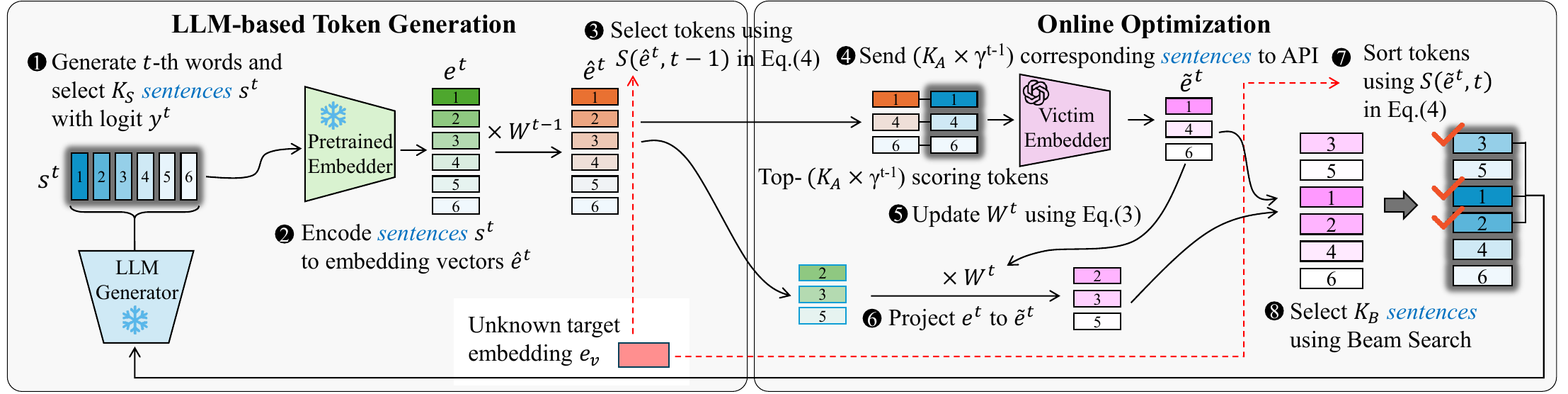}
    \caption{The recursive workflow of \sys{}: \ding{182} Generate diverse candidate tokens via LLM; \ding{183} Project local embeddings to the target space; \ding{184} Group candidates by confidence; \ding{185} Query the victim model for the top-$(K_A\times\gamma^{t-1})$ candidates; \ding{186} Update the alignment matrix online; \ding{187} Re-project non-queried candidates with the updated matrix; \ding{188} Re-score all candidates; and \ding{189} Select the best sequences via Beam Search for the next iteration.}
    \label{fig:architecture}
\end{figure*}

\subsection{Defense Mechanisms Against Inversion Attacks}
Despite the rapid advancement of inversion techniques~\cite{li2023sentence, huang2024transferable, chen2024against}, robust defense strategies remain underexplored. 
Early adversarial training methods~\cite{song2020information} proved ineffective in black-box scenarios where the attacker's model is unknown. 
To balance privacy and utility, \citet{morris2023text} proposed injecting Gaussian noise, though \citet{chen2024text} showed its efficacy diminishes in multilingual settings. 
Differential privacy (DP) offers a rigorous alternative; \citet{lyu2020differentially} applied DP during training, while \citet{du2023sanitizing} introduced a metric-based local differential privacy (LDP) mechanism for inference-time sanitization. 
In this work, we critically evaluate these mechanisms against our proposed \sys{} framework, highlighting the persistent challenges in securing embeddings against few-shot inversion threats.

%% file: 04method.tex
\section{Methodology}
\label{sec:method}

In this section, we propose \sys{}, a novel framework designed to recover original text from unknown target embeddings. 
As illustrated in Figure~\ref{fig:architecture}, \sys{} synergizes the linguistic priors of LLMs with a \textit{recursive online optimization} mechanism to achieve high reconstruction fidelity without accessing any auxiliary public corpora or training data.

\subsection{Problem Formulation}
We situate our work within a strict \textit{black-box} inversion scenario. 
The adversary aims to reconstruct a text sequence $x$ solely from its target embedding vector $e_v \in \mathbb{R}^d$, generated by an unknown victim model $\phi$. 
The adversary queries $\phi$ via an API but lacks access to model parameters $\theta$, gradients, or the training distribution. 

Crucially, unlike prior \textit{embedder alignment methods}~\cite{chen2025algen}, we assume a \textit{data-free} environment where the adversary possesses \textit{no leaked embedding-text pairs} to train a surrogate model or an adapter offline. The objective is to find a candidate $\hat{x}$ that minimizes the distance metric $\mathcal{D}$ using only the query budget:
\begin{equation}
    \hat{x}^* = \mathop{\mathrm{argmin}}_{\hat{x} \in \mathcal{X}} \mathcal{D}(\phi(\hat{x}), e_v)
\end{equation}

\subsection{Design Principles}
\label{ssec:design_principle}
In the absence of leaked reference data, existing inversion paradigms falter. \textit{Direct inversion methods}~\cite{morris2023text} fail due to data scarcity, while \textit{embedder alignment methods}~\cite{chen2025algen,huang2024teia} suffer from \textit{distribution shift}, where a static alignment learned on open corpora fails to generalize to the specific distribution of the target embedding. To address these challenges, \sys{} adopts two core principles:

\begin{itemize}[leftmargin=*, nosep]
    \item \textbf{Exploiting General-Purpose Linguistic Priors:} Instead of training a domain-specific decoder that risks overfitting to biased external data, we leverage the inherent, unbiased probability distributions of a pre-trained LLM to generate diverse candidates from scratch.
    \item \textbf{Targeted Online Optimization:} To eliminate the generalization gap caused by mismatched training distributions, we abandon static alignment. Instead, we propose \textit{targeted online optimization}, which dynamically optimizes the mapping matrix specifically for the single target embedding $e_v$ during inference, ensuring precise mapping without prior knowledge of the target domain.
\end{itemize}

\subsection{Iterative Token Generation}
\label{ssec:generation}
Following initialization, \sys{} iteratively reconstructs the source text token-by-token. This phase corresponds to steps \ding{182} and \ding{189} in Figure~\ref{fig:architecture}.

\paragraph{LLM-based Generation.}
At iteration $t$, given the context $s^{t-1}$ from the previous step (initially \texttt{[BOS]}), we leverage an open-source LLM (e.g., Qwen~\cite{yang2025qwen3}, Llama~\cite{touvron2023llama}) to predict the probability distribution over the vocabulary. The model generates logits $y^t$ for the next potential token. 
Note that to generate valid text, we restrict the target vocabulary to ASCII tokens.

\paragraph{Diversity-Aware Filtering.}
To prevent the candidates from collapsing into a single mode, we enforce diversity during selection. Instead of simply taking the top-$K$ tokens by probability, we select $K_S$ candidate tokens such that the pairwise cosine similarity between their embeddings remains below a predefined threshold $Th_w$ (\ding{182}). This ensures that the candidate set $s^t$ covers a broad range of semantic possibilities. Finally, we employ Beam Search to retain the top-$K_{B}$ most promising sentences for the next iteration (\ding{189}), repeating this process until the \texttt{[EOS]} token is generated or the maximum length $T$ is reached.

\subsection{Online Verification and Optimization}
\label{ssec:alignment_score}
To evaluate the generated candidates without querying the victim model for every sample (goals: accuracy \& query efficiency), we introduce an online verification and optimization module. 
This module dynamically maps our local embeddings to the victim's space and scores them reliably. 
This corresponds to steps \ding{183}--\ding{188} in Figure~\ref{fig:architecture}.

\paragraph{Online Projection Optimization.}
We compute a projector $W^t$ that projects the local embedding $e^t$ to the target vector space $\tilde{e}^t$ (\ding{183}). We formulate the update of $W^t$ as a Ridge Regression problem~\cite{mcdonald2009ridge}:
\begin{equation}
    \min_{W^{t}} \sum_{(e_{i}, \tilde{e}_{i})\in (E^{t}, \tilde{E}^{t})}
    \left\| e_{i}W^{t} - \tilde{e}_{i} \right\|^2
    + \lambda \left\| W^{t} \right\|^2,
    \label{eq:optim_w}
\end{equation}
where $E^t$ and $\tilde{E}^t$ denote the sets of local embeddings and ground-truth embeddings (queried from the victim) accumulated up to iteration $t$.
$\lambda$ is a regularization parameter.
To obtain $\tilde{E}^t$, we query the victim embedder with $K_{A} \times \gamma^{t-1}$ candidate sentences $s^{t}$ corresponding to the tokens selected in \ding{184}.
Here, $\gamma$ ($<1$) serves as an exponential decay factor that progressively reduces the number of queries sent to the victim embedder as iterations proceed (\ding{185}).
Note that in the first iteration, due to the absence of $W^{0}$, we bypass step \ding{184} after only generating $e^{1}$ in \ding{183} and exceptionally transmit $3 \times K_{A}$ queries to the victim model.
The closed-form solution is updated online as follows (\ding{186}):
\begin{equation}
    W^{t} = \left(E^{t \top}E^{t} + \lambda I\right)^{-1}
    E^{t \top}\tilde{E}^{t},
    \label{eq:calc_w}
\end{equation}
where $I$ denotes the identity matrix.
This allows us to re-project non-queried candidates using the latest $W^t$ (\ding{187}), ensuring accurate estimation with minimal API calls.

\paragraph{Confidence-Aware Scoring (Verification).}
To select the best candidates (groups \ding{184} \& sorting \ding{188}), we propose a hybrid scoring function $S(e_{i}, t)$ that serves as an online verification step, combining the LLM's logit priors with the projected embedding similarity:
\begin{equation}
    S(e_{i}, t) = \text{Z}(y_{i}) + \text{conf}_{t} \times \text{Z}(\text{cos}(e_{i}, e_{v})),
    \label{eq:score}
\end{equation}
where $\text{Z}(\cdot)$ denotes z-score normalization.
Here, $y$ represents the LLM's logits and serves to maintain grammatical consistency by capturing complex contextual dependencies.
Complementarily, the cosine similarity term $\text{cos}(e_{i}, e_{v})$ quantifies the semantic similarity between the embedding vector $e_{i}$ and the target embedding $e_{v}$.
Crucially, we introduce a dynamic confidence term $\text{conf}_t$ to weight the embedding similarity based on the reliability of the current projection $W^{t-1}$:
\begin{equation}
    \text{conf}_{t} = \frac{1}{|\textbf{E}^{t}|}\sum_{i\in \textbf{E}^{t}}\text{cos}(e^{t}_{i}W^{t-1}, \tilde{e}^{t}_{i}),
    \label{eq:confidence}
\end{equation}
where $\textbf{E}^{t}$ indicates the set of token indices sent to the API at the $t$-th iteration, $|\textbf{E}^{t}|$ represents the size of the index set.
This term ensures that the model relies more on the embedding similarity as the projection matrix $W$ becomes more accurate (higher $\text{conf}_t$), minimizing the risk of divergence in the early stages.
Note that for the initial iteration ($t=1$), since $W^{0}$ is undefined, $\text{conf}_{1} = 0.7\cdot \frac{1}{|\textbf{E}^{1}|}\sum_{i\in \textbf{E}^{1}}\text{cos}(e^{1}_{i}W^{1}, \tilde{e}^{1}_{i})$.

%% file: 05evaluation.tex
\section{Evaluation}
\label{sec:eval}

In this section, we evaluate the efficacy of \sys{} in reconstructing the original text, comparing it against SOTA textual embedding inversion attacks (e.g., Vec2Text, ALGEN, TEIA).
All experiments are conducted on a computing environment equipped with two NVIDIA RTX 5090 GPUs and an Intel Xeon 6530 processor.

\subsection{Experimental Setting}
\label{subsec:setting}

\begin{table*}[]
\centering
\caption{Performance comparison between \sys{} and SOTA methods.}
\label{tab:main_exp}
\resizebox{2\columnwidth}{!}{
\small
\begin{tblr}{
  colspec = {c|c|ccccc|ccccc},
  cell{19-22}{1-12} = {bg=gray!15},
  rowsep = 0pt,
}
\hline
\SetCell[r=2]{m}{Method}   & \SetCell[r=2]{m}{Victim model}  & \SetCell[c=5]{c}{MS MARCO} & & & & & \SetCell[c=5]{c}{PubMed} & & & &         \\ 
                           &                          & BLEU-1 & BLEU-2 & ROUGE-L & ROUGE-1 & COS   & BLEU-1 & BLEU-2 & ROUGE-L & ROUGE-1 & COS  \\\hline
\SetCell[r=4]{m}{Vec2Text\\w/o Corrector} 
                           & GTR-Base                 & 12.65  & 2.49   & 9.39    & 11.29   & 0.1703 & 11.34  & 2.30   & 8.30    & 10.06  & 0.1706  \\
                           & Qwen3-Embedding-0.6B     & 12.53  & 2.64   & 8.40    & 10.52   & 0.5166 & 11.39  & 2.48   & 8.41    & 10.50  & 0.4961  \\
                           & OPENAI(3-small)          & 12.65  & 2.71   & 8.59    & 10.85   & 0.1214 & 10.97  & 2.47   & 8.55    & 10.23  & 0.1215  \\
                           & OPENAI(3-large)          & 12.64  & 2.55   & 9.23    & 11.16   & 0.1043 & 11.15  & 1.99   & 8.09    & 9.53   & 0.1119  \\\hline
\SetCell[r=4]{m}{Vec2Text\\w/ Corrector}      
                           & GTR-Base                 & 12.72  & 2.56   & 9.20    & 11.16   & 0.1664 & 11.31  & 2.25   & 8.12    & 9.86   & 0.1768  \\
                           & Qwen3-Embedding-0.6B     & 12.82  & 2.60   & 8.00    & 9.94    & 0.4976 & 11.19  & 2.40   & 7.89    & 9.95   & 0.4841  \\
                           & OPENAI(3-small)          & 12.66  & 2.83   & 8.33    & 10.53   & 0.1207 & 11.06  & 2.49   & 8.46    & 10.13  & 0.1186  \\
                           & OPENAI(3-large)          & 12.54  & 2.44   & 8.91    & 10.85   & 0.1017 & 11.05  & 1.99   & 7.99    & 9.43   & 0.1109  \\\hline
\SetCell[r=4]{m}{TEIA}     & GTR-Base                 & 4.88   & 0.96   & 5.62    & 6.65    & 0.0182 & 4.97   & 0.97   & 6.36    & 7.62   & 0.0296  \\
                           & Qwen3-Embedding-0.6B     & 5.20   & 0.92   & 6.06    & 7.25    & 0.0220 & 4.16   & 0.89   & 5.22    & 5.91   & 0.0314  \\
                           & OPENAI(3-small)          & 4.16   & 0.85   & 5.46    & 6.59    & 0.0116 & 5.47   & 0.97   & 6.68    & 7.83   & 0.0274  \\
                           & OPENAI(3-large)          & 5.30   & 1.01   & 6.00    & 7.00    & 0.0180 & 3.73   & 0.86   & 5.25    & 6.17   & 0.0270  \\\hline
\SetCell[r=4]{m}{ALGEN}
                           & GTR-Base                 & 20.28  & 2.35   & 14.16   & 18.24   & 0.5574 & 19.52  & 1.42   & 13.76   & 17.70  & 0.5538  \\
                           & Qwen3-Embedding-0.6B     & 22.40  & 2.63   & 15.65   & 21.06   & 0.4022 & 22.37  & 2.35   & 14.96   & 19.73  & 0.4188  \\
                           & OPENAI(3-small)          & 19.85  & 1.87   & 13.69   & 17.96   & 0.2524 & 19.87  & 1.87   & 14.19   & 18.33  & 0.2478  \\
                           & OPENAI(3-large)          & 21.83  & 2.38   & 14.79   & 20.44   & 0.2483 & 21.27  & 1.97   & 13.94   & 18.25  & 0.2506  \\\hline
\SetCell[r=4]{m}{\sys{}}
                           & GTR-Base                 & 24.95  & 10.95  & 18.73   & 26.44   & 0.7514 & 21.34  & 8.38   & 16.81   & 22.00  & 0.7174  \\
                           & Qwen3-Embedding-0.6B     & 29.29  & 13.45  & 22.55   & 32.22   & 0.7323 & 25.29  & 10.87  & 19.92   & 27.59  & 0.7096  \\
                           & OPENAI(3-small)          & 34.44  & 17.14  & 25.55   & 37.85   & 0.6983 & 35.42  & 17.97  & 25.64   & 36.39  & 0.6859  \\
                           & OPENAI(3-large)          & 32.43  & 16.42  & 26.08   & 37.52   & 0.6634 & 33.76  & 17.17  & 24.61   & 35.31  & 0.6403  \\\hline
\end{tblr}
}
\end{table*}

\paragraph{Models.} 

For LLM-based token generation, \sys{} utilizes the pretrained Qwen3-0.6B model~\cite{yang2025qwen3} as the LLM generator, while employing all-mpnet-base-v2~\cite{reimers2019sbert} as the attacker's local embedder.
To evaluate \sys{} across various victim models, we utilize GTR-Base~\cite{ni2022large} and Qwen3-Embedding-0.6B~\cite{zhang2025qwen3} as open-source victim embedding models, and OpenAI's Text-Embedding-3-small (3-small)~\cite{SmallOpenai} and Text-Embedding-3-large (3-large)~\cite{LargeOpenai} as closed-source victim embedding models.
Detailed descriptions of each embedder are provided in Section B of supplementary material.

\paragraph{Datasets.}

To evaluate the reconstruction capabilities of \sys{} in terms of cross-domain generalization, we employ two distinct datasets: MS MARCO~\cite{bajaj2016ms} for the general domain and PubMed~\cite{cohan-etal-2018-discourse} for the medical-specific domain.
MS MARCO is a large-scale dataset widely used in information retrieval, covering a wide range of general topics.
PubMed comprises over 36 million citations and abstracts from biomedical literature, representing highly specialized technical texts.
Following prior work~\cite{chen2025algen}, we evaluate the reconstruction performance on 200 sampled texts from each dataset.

\paragraph{Metrics.} 

To evaluate the inversion performance, we use five metrics: BLEU-1, BLEU-2, ROUGE-L, ROUGE-1, and COS.
BLEU-1 and BLEU-2~\cite{papineni2002bleu} focus on n-gram precision to evaluate word matching between the original and reconstructed texts.
In contrast, ROUGE-L and ROUGE-1~\cite{lin2004rouge} emphasize recall, specifically measuring the longest common subsequence (LCS) and lexical overlap to assess sentence-level structural similarity.
Additionally, we utilize COS to quantify the semantic similarity between the two texts.
Specifically, COS calculates the cosine similarity between the target embedding vector and the embedding vector obtained by feeding the reconstructed sentence into the victim model.

\paragraph{Considered Approaches.}

We evaluate \sys{} against three SOTA methods: Vec2Text~\cite{morris2023text}, TEIA~\cite{huang2024teia}, and ALGEN~\cite{chen2025algen}.
Unlike \sys{}, which reconstructs target text under a strict \textit{black-box} setting, existing approaches necessitate offline decoder training or alignment using the same-domain data with a target embedding vector.
To ensure a fair comparison, we assume that existing approaches collect training text-embedding pairs by querying the victim model with 1,000 texts from the MultiHPLT English dataset~\cite{de2024new}, which is distinct from the evaluation domain.
We limit the number of queries to 1,000, following the experimental settings of ALGEN~\cite{chen2025algen}.
The specific training and reconstruction strategies for each approach are as follows:

\begin{itemize}
    \item \textbf{Vec2Text:} The decoders (Base and Corrector), both based on T5-base (235M), are trained on the 1,000 embedding-text pairs. During the reconstruction phase, Vec2Text employs 50 iterations and beam search with a beam size of 8, coupled with the corrector, which iteratively queries the victim model to refine the generated text. In our experiments, we report results for both without Corrector and with Corrector. % \dhcmt{iteration 횟수 추가하기}
    \item \textbf{TEIA:} It utilizes DialoGPT-small (117M) to train a surrogate embedding model (specifically, an adapter) utilizing the 1,000 embedding-text pairs.
    \item \textbf{ALGEN:} The decoder of ALGEN, based on Flan-T5-small (80M), is trained on 150k pairs consisting of MultiHPLT and embeddings from an open-source embedder. The 1,000 text-embedding pairs obtained from the victim model are subsequently utilized for the alignment phase.
\end{itemize}

\paragraph{Hyperparamters.}
For text reconstruction, \sys{} employs beam search with a beam size of 10.
We set $K_S=1000$, $K_{A}=50$, $\gamma=0.8$, $Th_{w}=0.9$, $T=32$ and $\lambda=0.1$.
For the generation constraints, we apply a logit penalty of -5 to non-alphabetic tokens at the first iteration.

\subsection{Comparison to SOTA Methods}

\paragraph{Inversion Attack Performance.}

Table~\ref{tab:main_exp} shows the text reconstruction performance of each approach on the MS MARCO and PubMed datasets.
Vec2Text and TEIA exhibit substantially degraded performance across all datasets and victim models, primarily due to the severe scarcity of data available for offline training.
In particular, for Vec2Text, the performance gains from utilizing online API queries during reconstruction are marginal, as the Corrector is insufficiently trained to leverage the feedback effectively.
Although ALGEN employs a decoder trained on large-scale open datasets, it suffers from significant performance drops due to the domain discrepancy between the alignment data and the target.
These results demonstrate the vulnerability of existing methods in cross-domain scenarios, where the domain of the offline prepared dataset differs from that of the actual dataset to be reconstructed.

In contrast, \sys{} consistently outperforms existing methods on all metrics regardless of the victim model or dataset.
For instance, on the MS MARCO dataset and the OpenAI (3-large) victim model, \sys{} achieves a ROUGE-L score of 26.08, approximately 1.8 times higher than ALGEN's ROUGE-L score of 14.79.
Moreover, \sys{} achieves a BLEU-2 score of 16.42, which is approximately 6.4 times higher than the 2.55 achieved by Vec2Text (w/o Corrector).
These results suggest that by jointly considering syntactic and semantic information as defined in Eq.~\eqref{eq:score}, \sys{} successfully reconstructs not only single words but also continuous word sequences.

\begin{table}[]
\centering
\caption{Analysis of the number of queried sentences and tokens.}
\label{tab:num_query}
\resizebox{0.73\columnwidth}{!}{
\begin{tabular}{c|cc|cc}
\hline
\multirow{2}{*}{Method}       & \multicolumn{2}{c|}{Sentence}                              & \multicolumn{2}{c}{Token ($\times10^3$)} \\
                              & \multicolumn{1}{c}{Offline}  & \multicolumn{1}{c|}{Online} & \multicolumn{1}{c}{Offline}    & Online  \\ \hline
Vec2Text                      & 1000                         & 3144                        & 28.45                          & 82.60     \\
ALGEN                         & 1000                         & 0                           & 28.45                          & 0       \\
TEIA                          & 1000                         & 0                           & 30.15                          & 0       \\
\sys{}                        & 0                            & 2180                        & 0                              & 13.88   \\ \hline
\end{tabular}
}
\end{table}

\paragraph{API Query Cost.}

Table~\ref{tab:num_query} shows the API query costs for each method, measured by the number of sentences and tokens transmitted to the OPENAI(3-small) victim model during both offline and online phases.
The number of queried tokens was calculated based on the MultiHPLT dataset for the offline phase and the PubMed dataset for the online phase.
For the offline training, Vec2Text, ALGEN, and TEIA require querying the victim model with 1,000 sentences to construct their training or alignment datasets.
Since each approach employs a different internal tokenizer, the number of resulting tokens varies: Vec2Text and ALGEN send 28.45k tokens, and TEIA sends 30.15k tokens (detailed in Section A of supplementary material).
In contrast, \sys{} requires no offline training, incurring zero query costs (both in sentences and tokens).

In the online reconstruction phase, ALGEN, TEIA, and Vec2Text (without Corrector) do not perform query-based refinement, resulting in zero online query expenditure.
Conversely, Vec2Text (with Corrector) and \sys{} actively query the victim model, transmitting an average of 3,144 and 2,180 sentences, respectively.
Notably, \sys{} is significantly more token-efficient, consuming only 13.88k tokens on average, compared to Vec2Text, which consumes 82.60k tokens.
This efficiency is attributed to the fact that \sys{} utilizes short token sequences in early iterations and progressively reduces the number of queried sentences via the decay in $\gamma$ in later iterations.

\begin{figure}[]
    \centering
    \includegraphics[width=0.97\linewidth]{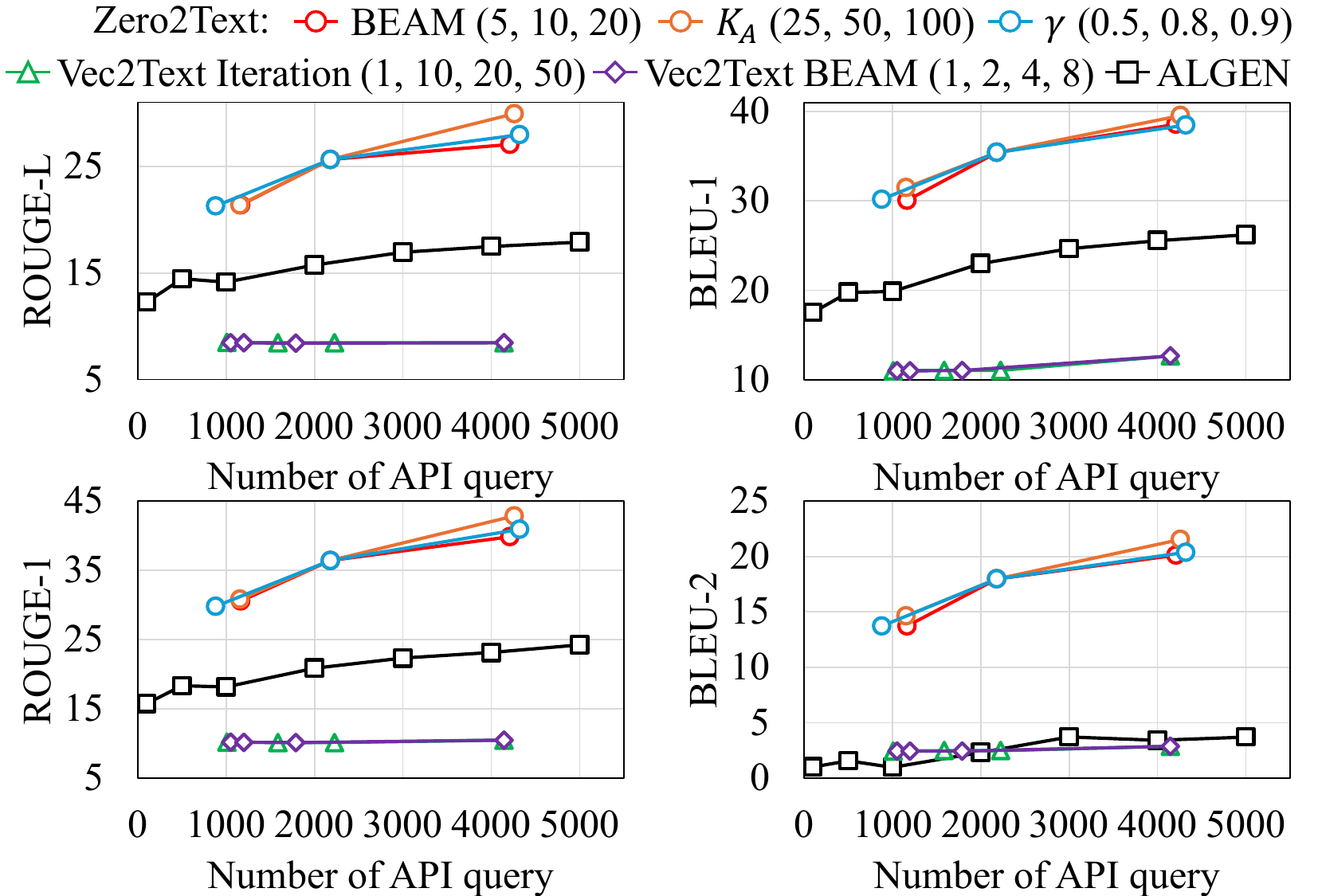}
    \caption{Comparison inversion performance between Vec2Text, ALGEN and \sys{} depending on the number of queries.}
    \label{fig:number_of_query}
\end{figure}

\begin{figure*}[]
    \centering
    \includegraphics[width=0.95\linewidth]{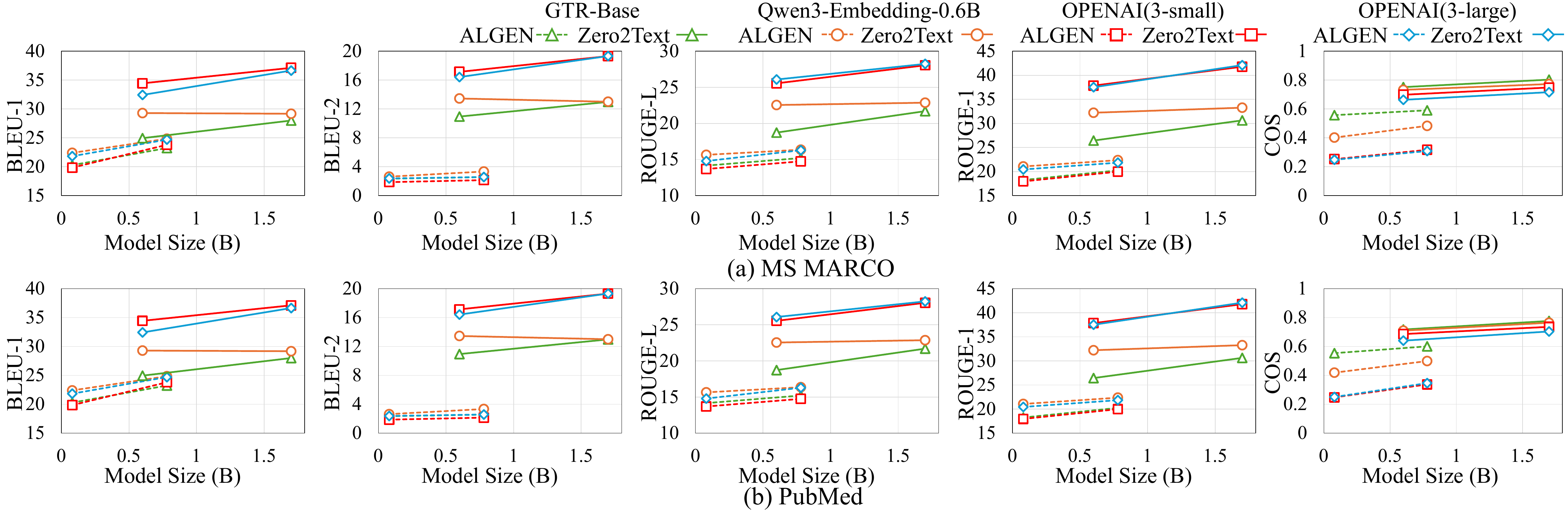}
    \caption{Scalability analysis of inversion attack performance between ALGEN and Zero2Text.}
    \label{fig:model_size}
\end{figure*}

\begin{table*}[]
\centering
\caption{Impact of \sys{}'s pretrained embedder choice on reconstruction performance.}
\label{tab:attacker_embedder}
\resizebox{2\columnwidth}{!}{
\small
\begin{tblr}{
  colspec = {c|c|ccccc|ccccc},
  rowsep = 0pt,
}
\hline
\SetCell[r=2]{m}{Attacker embedder}   & \SetCell[r=2]{m}{Victim model}  & \SetCell[c=5]{c}{MS MARCO} & & & & & \SetCell[c=5]{c}{PubMed} & & &  &   \\ 
                           &                          & BLEU-1 & BLEU-2 & ROUGE-L & ROUGE-1 & COS    & BLEU-1 & BLEU-2 & ROUGE-L & ROUGE-1 & COS  \\\hline
\SetCell[r=4]{m}{mE5-Large-Instruct} 
                           & GTR-Base                 & 33.17  & 15.40  & 23.61   & 35.44   & 0.8660       & 29.31  & 12.79 & 20.44    & 28.72  & 0.8356      \\
                           & Qwen3-Embedding-0.6B     & 33.77  & 16.07  & 25.22   & 38.47   & 0.8187       & 29.63  & 13.72 & 20.55    & 32.39  & 0.8288      \\
                           & OPENAI(3-small)          & 40.36  & 20.94  & 29.30   & 44.18   & 0.7736       & 37.11  & 19.29 & 26.26    & 38.99  & 0.7738      \\
                           & OPENAI(3-large)          & 37.76  & 19.68  & 29.16   & 43.20   & 0.7470       & 38.35  & 19.09 & 24.79    & 38.35  & 0.7440      \\\hline
\end{tblr}
}
\end{table*}

\subsection{Varying Number of Queries}

To investigate the trade-off between API query costs and reconstruction performance, we conducted an experiment by varying the parameters that control the query budget.
Figure~\ref{fig:number_of_query} visualizes the reconstruction performance of \sys{}, ALGEN, and Vec2Text depending on the number of queries.
The y-axis respectively represents the reconstruction performance metrics (ROUGE-L, ROUGE-1, BLEU-1, and BLEU-2), while the x-axis denotes the total number of queries in both offline and online phases.
For \sys{}, we adjust the online query volume by adjusting the beam size, $K_A$, and $\gamma$.
Similarly, for Vec2Text, the query budget is controlled via the beam size and the number of Corrector iterations.
In the case of ALGEN, which does not utilize online queries, we vary the offline query budget over the set $\{100, 500, 1000, 2000, 3000, 4000, 5000\}$.

As shown in the Figure~\ref{fig:number_of_query}, increasing the beam size, $K_A$, and $\gamma$ in \sys{} leads to a higher query count, which in turn monotonically improves the inversion performance.
ALGEN exhibits a similar trend where performance improves with increased offline queries.
However, the rate of improvement gradually diminishes, showing a saturation trend.
In contrast, Vec2Text shows negligible performance gains even with a substantial increase in the number of queries.

\begin{table}[]
\centering
\caption{Comparison of alignment performance on the PubMed dataset.}
\label{tab:alignment}
\resizebox{0.73\columnwidth}{!}{
\begin{tabular}{c|c|c|c}
\hline
Method                  & Alignment                 & Victim model          & COS      \\\hline
\multirow{2}{*}{TEIA}   & \multirow{2}{*}{Offline}  & OPENAI(3-small)       & 0.0039   \\
                        &                           & OPENAI(3-large)       & 0.0056   \\\hline
\multirow{2}{*}{ALGEN}  & \multirow{2}{*}{Offline}  & OPENAI(3-small)       & 0.3229   \\
                        &                           & OPENAI(3-large)       & 0.2848   \\\hline
\multirow{2}{*}{\sys{}} & \multirow{2}{*}{Online}   & OPENAI(3-small)       & 0.6063   \\
                        &                           & OPENAI(3-large)       & 0.5447   \\\hline

\end{tabular}
}
\end{table}

\subsection{Alignment Performance}

Table~\ref{tab:alignment} compares the performance of the embedder alignment methods (TEIA, ALGEN) and \sys{}.
To evaluate this, we measure the cosine similarity between the target embedding vector (obtained from the victim model) and the projected embedding vector generated by processing the ground-truth text through each method's aligner.
TEIA trains a simple MLP-based aligner offline.
However, under the strict \textit{black-box} scenario, the scarcity of offline training data prevents the aligner from effectively mapping vectors into the victim model's embedding space, resulting in a performance drop.
Similarly, ALGEN fails to achieve sufficient alignment optimization due to its limited generalization capability across domains.
In contrast, \sys{} performs instance-level optimization of the aligner during the online phase, thereby achieving a substantial cosine similarity of over 0.5.
These results highlight the crucial role of online alignment in achieving high reconstruction performance within strict \textit{black-box} scenarios.

\subsection{Model Scalability}

To investigate the impact of model capacity on reconstruction performance, we evaluate both ALGEN and \sys{} across different model sizes.
Figure~\ref{fig:model_size} visualizes the reconstruction results on the MS MARCO and PubMed datasets across four victim models.
The x-axis represents the model size (in billions of parameters), and the y-axis denotes the reconstruction performance.
We compare \sys{}, employing QWEN3-0.6B and QWEN3-1.7B as the LLM Generator, against ALGEN utilizing Flan-T5-small (0.08B) and Flan-T5-large (0.78B).
As shown in Figure~\ref{fig:model_size}, reconstruction performance generally improves with increased model size.
However, even with the larger model, ALGEN consistently yields lower reconstruction performance compared to \sys{}.
For instance, on the MS MARCO dataset against the OpenAI (3-large) victim model, \sys{} achieves a ROUGE-L score approximately 1.6 times higher than that of ALGEN.
This suggests that \sys{} is highly parameter-efficient, maintaining superior reconstruction performance even when utilizing smaller models than ALGEN.
Furthermore, while ALGEN incurs additional training costs to utilize larger models, \sys{} has the advantage of employing larger pretrained models without incurring any extra training overhead.

\subsection{Attacker Embedder}

To assess the robustness of \sys{} with respect to the pretrained embedder configuration, we evaluate the reconstruction performance when replacing the default embedder with mE5-Large-Instruct~\cite{wang2024multilingual} during the LLM-based token generation phase.
Table~\ref{tab:attacker_embedder} presents the reconstruction results of \sys{} on the MS MARCO and PubMed datasets.
As shown in the Table~\ref{tab:attacker_embedder}, \sys{} maintains high reconstruction performance across both datasets and victim models.
For instance, against the GTR-Base victim model on the MS MARCO dataset, \sys{} achieves a ROUGE-1 score of 35.44.
Since \sys{} allows for the integration of the attacker embedder without any additional training costs, these results suggest that \sys{} offers the flexibility to employ a diverse range of embedders, potentially further enhancing reconstruction performance.

\begin{table}[]
\centering
\caption{Performance comparison of \sys{} with variaus defense method.}
\label{tab:defense}
\resizebox{0.98\columnwidth}{!}{
\begin{tabular}{c|c|c|ccccc}
\hline
\multirow{2}{*}{Method}     &\multirow{2}{*}{Defense}   & \multirow{2}{*}{$\epsilon/d$} & \multicolumn{5}{c}{PubMed}                    \\ 
                            &                           &       & BLEU-1 & BLEU-2 & ROUGE-L & ROUGE-1 & COS    \\ \hline
\multirow{13}{*}{ALGEN}     &\multirow{1}{*}{None}      & -     & 19.87  & 1.87   & 14.19   & 18.33   & 0.2478    \\ \cline{2-8}
                            &\multirow{1}{*}{Random}    & -     & 12.61  & 0.52   & 8.85    & 10.32   & 0.0851    \\ \cline{2-8}
%                            &\multirow{1}{*}{WET}       & -     & 20.01  & 1.88   & 14.32   & 19.10   & 0.3090    \\ \cline{2-8}
                            &\multirow{5}{*}{LapMech}   & 0.25  & 13.82  & 0.69   & 9.60    & 12.02   & 0.1003    \\
                            &                           & 0.5   & 15.03  & 0.86   & 10.63   & 12.94   & 0.1409    \\
                            &                           & 1     & 15.73  & 0.80   & 11.23   & 13.81   & 0.1767    \\
                            &                           & 2     & 18.03  & 1.33   & 12.48   & 15.75   & 0.2202    \\
                            &                           & 4     & 18.62  & 1.61   & 13.69   & 17.50   & 0.2432    \\ \cline{2-8}
                            &\multirow{5}{*}{PurMech}   & 0.25  & 11.97  & 0.50   & 7.95    & 9.28    & 0.0963    \\
                            &                           & 0.5   & 14.27  & 0.55   & 10.07   & 12.02   & 0.1403    \\
                            &                           & 1     & 15.98  & 0.93   & 12.07   & 14.92   & 0.1990    \\
                            &                           & 2     & 18.07  & 1.34   & 13.21   & 16.23   & 0.2250    \\
                            &                           & 4     & 19.34  & 1.88   & 13.81   & 18.09   & 0.2470    \\\hline
\multirow{13}{*}{\sys{}}    &\multirow{1}{*}{None}      & -     & 35.42  & 17.97  & 25.64   & 36.39   & 0.6859    \\ \cline{2-8}
                            &\multirow{1}{*}{Random}    & -     & 15.57  & 3.49   & 9.67    & 11.57   & 0.1463    \\ \cline{2-8}
%                            &\multirow{1}{*}{WET}       & -     &        &        &         &         &           \\ \cline{2-8}
                            &\multirow{5}{*}{LapMech}   & 0.25  & 20.36  & 6.99   & 13.75   & 17.98   & 0.1893    \\
                            &                           & 0.5   & 27.37  & 11.69  & 19.27   & 26.56   & 0.3060    \\
                            &                           & 1     & 32.45  & 15.97  & 22.93   & 32.35   & 0.4816    \\
                            &                           & 2     & 34.30  & 17.29  & 24.14   & 34.29   & 0.6094    \\
                            &                           & 4     & 34.04  & 17.20  & 25.11   & 35.29   & 0.6649    \\ \cline{2-8}
                            &\multirow{5}{*}{PurMech}   & 0.25  & 19.93  & 6.26   & 13.50   & 17.98   & 0.1894    \\
                            &                           & 0.5   & 27.73  & 11.13  & 18.84   & 26.64   & 0.3016    \\
                            &                           & 1     & 32.88  & 15.47  & 22.99   & 32.53   & 0.4828    \\
                            &                           & 2     & 33.19  & 16.14  & 23.93   & 33.81   & 0.6036    \\
                            &                           & 4     & 34.65  & 17.32  & 25.20   & 36.02   & 0.6657    \\\hline
\end{tabular}
}
\end{table}

\subsection{Defense}

\paragraph{Considerable Defense Approaches.} 

To assess the robustness of \sys{} against embedding protection mechanisms, we evaluate its reconstruction performance under distinct defense strategies: Random noise and Local Differential Privacy (LDP).
The random noise strategy serves as a baseline by injecting random perturbations directly into the target embedding.
For LDP, we employ the Normalized Planar Laplace (LapMech)~\cite{dwork2014lapmech} and Purkayastha Mechanism (PurMech)~\cite{du2023purmech} to ensure metric-LDP.
In our experiments, the privacy budget is configured as $\epsilon/d \in \{0.25, 0.5, 1, 2, 4\}$, normalized by the embedding dimension $d$.
Note that a smaller $\epsilon/d$ corresponds to stronger noise injection.

\paragraph{Defense Results.} 

Table~\ref{tab:defense} presents the reconstruction performance of ALGEN and \sys{} when subjected to these defense mechanisms.
While both methods experience a degradation in performance as defenses are applied, \sys{} exhibits remarkable robustness.
Even under defense mechanisms, \sys{} maintains a reconstruction performance comparable to that of ALGEN without any defense.
For instance, when PurMech is applied with a high noise intensity ($\epsilon/d=0.25$), \sys{} achieves a BLEU-1 score of 20.36 and a ROUGE-L score of 13.75. These results are competitive with the performance of undefended ALGEN, which achieves 19.87 and 14.19, respectively.

\subsection{Qualitative Analysis}

\begin{table}[]
\caption{Qualitative analysis of inversion results on the MS MARCO dataset against the Qwen3-Embedding-0.6B victim model. \textcolor{blue}{\textbf{Blue}} and \textcolor{red}{\textbf{Red}} denote reconstructed proper nouns and continuous word sequences matched with the original text, respectively.}
\label{tab:compare_text}
\resizebox{0.95\columnwidth}{!}{
\begin{tabular}{c| >{\arraybackslash}m{7cm}}
\hline\hline            
Original text    &  A protocol for \textcolor{red}{\textbf{building automation}}. \textcolor{blue}{\textbf{BACnet}} is a data \textcolor{red}{\textbf{communication protocol}} mainly \textcolor{red}{\textbf{used in}} the \textcolor{red}{\textbf{building automation}} and HVAC industry (Heating Ventilation and Air     \\ \hline
TEIA             &  If you're wondering how to subscribe to the Memo Preview Program for high school kids to scout for the next season, go to Memo.       \\ \hline
Vec2Text         &  “ section has transition history system has unique, and it don have a special unique in common. and. We example, we planet has itsa unique       \\ \hline
ALGEN            &  In addition to UL’s CIO-CNC, the Intelligent Storage Unit is a modular solution for building data structures. UL provides       \\ \hline
\sys{}           &  This is a \textcolor{blue}{\textbf{BACnet}} protocol, \textcolor{red}{\textbf{used in}} \textcolor{red}{\textbf{building automation}} systems for communication between devices.  It is the most widely used \textcolor{red}{\textbf{communication protocol}} in the building industry. \\ \hline \hline
\end{tabular}
}
\end{table}

Table~\ref{tab:compare_text} presents a qualitative comparison between the original texts and the texts reconstructed by each method.
While existing SOTA methods often construct isolated words, they consistently fail to accurately reconstruct proper nouns or consecutive word sequences.
In contrast, as shown in Table~\ref{tab:compare_text}, \sys{} successfully reconstruct complex elements, including proper nouns and continuous word sequences.
Additional qualitative examples are provided in Section D of supplementary material.

%% file: 06conclusion.tex
\section{Conclusion}
\label{sec:conclusion}

In this paper, we addressed the critical limitation of existing inversion methods: their poor cross-domain generalization under strict \textit{black-box} scenarios.
To this end, we propose \sys{}, a novel training-free inversion attack method that integrates LLM-based token generation with a new instance-level online alignment strategy.
By eliminating the need for training or prior domain knowledge, \sys{} effectively reconstructs original texts even in strict \textit{black-box} scenarios.
Our experiments on the MS MARCO dataset against the OpenAI (3-large) victim model validate the effectiveness of our approach, achieving 1.8$\times$ higher ROUGE-L and 6.4$\times$ higher BLEU-2 scores compared to SOTA methods.
These results expose critical vulnerabilities, necessitating robust defenses against such generalization-capable attacks.